%% file: main.tex
\documentclass{article}

\usepackage{PRIMEarxiv}

\usepackage[utf8]{inputenc} 
\usepackage[T1]{fontenc}    
\usepackage{hyperref}       
\usepackage{url}            
\usepackage{booktabs}       
\usepackage{amsfonts}       
\usepackage{nicefrac}       
\usepackage{microtype}      
\usepackage{lipsum}
\usepackage{fancyhdr}       
\usepackage{graphicx}       
\graphicspath{{media/}}     
\usepackage{graphicx} 
\usepackage{natbib}
\usepackage{amsmath}
\usepackage{url}
\usepackage{subcaption}

\global\long\def\bfX{\mathbf{X}}

\global\long\def\ep{\mathbb{E}}
\usepackage{amsfonts}

\global\long\def\ep{\mathbb{E}}

\global\long\def\cn{\mathcal{N}}

\global\long\def\bfX{\mathbf{X}}

\global\long\def\ep{\mathbb{E}}
\global\long\def\cn{\mathcal{N}}

\global\long\def\ep{\mathbb{E}}

\global\long\def\cn{\mathcal{N}}
\newtheorem{definition}{Definition}[section]

\newcommand\independent{\protect\mathpalette{\protect\independenT}{\perp}}
\def\independenT#1#2{\mathrel{\rlap{$#1#2$}\mkern2mu{#1#2}}}
\usepackage{tikz}
\usetikzlibrary{shapes,decorations,arrows,calc,arrows.meta,fit,positioning}
\tikzset{
    -Latex,auto,node distance =0.6 cm and 0.6 cm, thick, line width = 1.5,
    state/.style ={circle, draw, thick, minimum width = 0.8 cm, line width=1pt},
    point/.style = {circle, draw, inner sep=0.04cm,fill,node contents={}},
    bidirected/.style={Latex-Latex,dashed},
    el/.style = {inner sep=2pt, align=left, sloped},
    every picture/.style={line width=1pt}
}
\newcommand{\cut}[1]{}
\title{Disparate Effect Of Missing Mediators On Transportability of Causal Effects
}

\author{
  Vishwali Mhasawade\\
  New York University\\
  \texttt{vishwalim@nyu.edu} \\
   \And
 Rumi Chunara\\
 New York University\\
  \texttt{rumi.chunara@nyu.edu} \\
}

\begin{document}
\maketitle

\begin{abstract}
Transported mediation effects provide an avenue to understand how upstream interventions (such as improved neighborhood conditions like green spaces) would work differently when applied to different populations as a result of factors that mediate the effects. However, when mediators are missing in the population where the effect is to be transported, these estimates could be biased. We study this issue of missing mediators, motivated by challenges in public health, wherein mediators can be missing, not at random. We propose a sensitivity analysis framework that quantifies the impact of missing mediator data on transported mediation effects. This framework enables us to identify the settings under which the conditional transported mediation effect is rendered insignificant for the subgroup with missing mediator data. Specifically, we provide the bounds on the transported mediation effect as a function of missingness. We then apply the framework to longitudinal data from the Moving to Opportunity Study, a large-scale housing voucher experiment, to quantify the effect of missing mediators on transport effect estimates of voucher receipt, an upstream intervention on living location, in childhood on subsequent risk of mental health or substance use disorder mediated through parental health across sites. Our findings provide a tangible understanding of how much missing data can be withstood for unbiased effect estimates.
\end{abstract}


\input{introduction}

\input{related_work}
\input{methods}

\input{experiments}
\input{discussion}

\bibliographystyle{plainnat}  
\bibliography{ref}

\end{document}

%% file: introduction.tex
\section{Introduction}
There is a growing understanding of the importance of multi-level causes and determinants of health \citep{diez2000multilevel}. For example, factors at the neighborhood level, such as available healthy food resources, affect health outcomes such as hypertension directly as well as indirectly through individual behaviors like food consumption/diet \citep{motamedi2021diet,stevenson2023neighborhood}. 
This perspective opens the possibility of intervening at different levels; for example, increasing green space in a neighborhood or a behavioral intervention on physical activity both for impacting cardiovascular risk \citep{dalton2020residential,stevenson2023neighborhood}.
Notably, when estimating the causal effect of interventions at distal (e.g., neighborhood) levels, the mechanism through which such a factor acts is often through individual-level factors like physical activity or diet behaviors \citep{mohnen2012health}.
Therefore, when considering distal causes, the inclusion of mediators, covariates that mediate the relationship between distal causes and outcomes, becomes essential for assessing, comparing, and transporting treatment effects. 
Accordingly, here we focus on questions with treatments at a distal level, in the context of the transportability of causal effects. 
We present a method for quantifying the effect of missing mediators on the estimated transported causal effect and study the relationship of missing mediators with causal effect measurement. Specifically, we show how disparate impact can arise on disadvantaged (e.g. minoritized or under-represented) versus advantaged subgroup populations based on differing proportions of missing mediator data.

We first highlight the challenges that missing mediators introduce in the setting of transporting causal effects. 
Transporting effects across populations where the source and target differ is a common task, especially when practitioners have access to observational data from multiple environments with potentially different populations. 
This is desirable when one wants to identify if the conclusions from a specific experiment apply to a different population, which is known as identifying the transportability of treatment effects or generalizing experimental results.
Approaches aimed at identifying the transportability of treatment effects across two environments are based on two main assumptions: 1) knowledge about the differences across environments, which are usually represented using selection nodes in a selection diagram, an augmented causal graph, and 2) complete data is observed across both the environments, i.e., there are no missing attributes in the target environment \citep{cinelli2021generalizing,bareinboim2012transportability}. 
However, it is difficult to obtain complete data from all the environments when assessing the transportability of causal effects due to different data collection policies, privacy concerns, etc.
Accordingly, here, we focus on the second assumption regarding observing complete data and study under what conditions transportability is identifiable when the assumption is violated.
Specifically, we focus on the setting where mediators are observed in the target environment but partially missing.  

The issue of missing mediators is of further importance, given that such data can be missing disparately.
Specifically, it has been shown that respondents with missing information may not be a random subset of population-based survey participants and may differ on other relevant sociodemographic characteristics such as neighborhood socioeconomic status (SES) \citep{kim2007potential,krieger2008race}. 
Furthermore, missing data about individual-level behaviors such as alcohol consumption is also linked to neighborhood SES \citep{sania2021k}. 
For example, pregnant women of low socioeconomic background are more likely to access antenatal care late in pregnancy, enroll late in research studies, and,
therefore, have more missing data early in pregnancy \citep{simkhada2008factors}. 
This is problematic as SES is an important determinant of drinking behavior during pregnancy \citep{skagerstrom2011predictors}. In sum, it is important to note that mediators can be missing in relation to the distal factors of interest as well.

To study this effect systematically, we present a causal view of the setting using a directed acyclic graph (DAG) and highlight the settings under which the causal effect can be identified under different missingness scenarios.
We adapt estimators for transporting causal effects under missing data and present a framework to evaluate the sensitivity of the estimated causal effect to missing mediator data.
Our findings highlight the need for robust sensitivity analysis of missing data, specifically missing mediators in the context of the transportability of causal effects. 
We demonstrate the effect of missing mediators on transported indirect treatment effect estimation from the Moving to Opportunity Study \citep{sanbonmatsu2011moving}, which has been used to study the causal effect of moving to a new neighborhood.

%% file: related_work.tex
\section{Previous work}
\noindent \textbf{Identification for transportability of causal effects.}
Identifying the transportability of causal effects across environments under selection bias has primarily focused on complete data observed in both the source and target environment \citep{pearl2011transportability,bareinboim2012transportability, bareinboim2013meta,bareinboim2014transportability}. 
However, in applications such as epidemiology and public health, there can be challenges due to the high resources required for data collection, resulting in the inability to observe all the covariates in the target environment.
For example, social determinants of health represent one such class of attributes that can be difficult to obtain \citep{braveman2014social,cantor2018integrating}. 
The main focus on the identifiability of transported causal effects has been on obtaining the various conditions under which an optimal adjustment set can be obtained, although restricted to the assumption that complete data is observed in the target environment \citep{henckel2019graphical,bareinboim2012transportability,shpitser2012validity}. 
In another line of work, selecting specific covariates to ensure that selection bias does not affect the population average treatment effect (PATE) has also been discussed \citep{egami2021covariate}. This relies on identifying a separating set of covariates, $\bfX$, such that $Y\independent S \mid \bfX$.
The optimal adjustment set, often also known as a separating set, is also used for designing robust prediction models under distribution shift \citep{subbaswamy2019preventing,singh2021fairness}. 
However, to estimate causal effects in the target environment by correcting for selection bias, such a separating set would need to be modified when certain covariates are unobserved.
Thus, although in the observational data setting, identification of causal effects for transport under selection bias is developed, the specific issue with missing mediators and selection bias in the target environment is largely unexplored.\\

\noindent \textbf{Estimating causal effects under selection bias.} 
For estimating causal effects in the target environment, it is important to quantify the bias between the environments. One standard approach is to quantify the bias between sample average treatment effect (SATE) and population average treatment effect (PATE) under selection bias assuming a linear model as SATE $-$ PATE = $b_{ax} \big[\frac{P(X=1)}{P(S=1) } (P(S=1\mid X=1) - P(S=1)) \big]$ where $b_{ax}$ denotes interaction coefficient between the treatment $a$ and the pre-treatment covariate $x$ in a linear model and $S=1$ represents the target environment \citep{seamans2021generalizability}.
Accordingly, when addressing the issue of transportability of causal effects under selection bias, accounting for the sampling probability of the study sample is pertinent to address the bias between the SATE and PATE.  
This idea has been explored in the context of inverse probability of selection weights \citep{cole2010generalizing}, inverse propensity score weighting \citep{robertson2021estimating}, and outcome regression methods \citep{robertson2021regression}.
However, these methods do not explicitly address missing values. 
Thus, even though there exist methods which can help address selection bias, work describing guarantees when covariates (including mediators, specifically) are missing or inconsistent in the target environment is lacking.\\

\noindent \textbf{Causal inference with missing covariates and selection bias.} 
Transporting causal effects when the same covariates are not measured completely in the study and the target population is challenging as there could be a covariate shift across environments and collecting data with no missingness is challenging.
Quantifying and correcting the bias requires prior information on what covariates, pre-treatment, confounders, colliders, moderators, treatment, or mediators are observed in the target environment.
The setting when the pre-treatment covariates are only measured in the source environment but not in the target environment has been studied in previous work \citep{andrews2019simple}. 
In this case, bias due to missingness and selection bias are both corrected using the target environment covariate means. Briefly, the causal estimates in the source environment are adjusted based on the mean of the covariates from the target environment as selection bias introduces differences between the source and target mean values. 
The assumption here is that even though the covariates are missing in the target population, their descriptive statistics are known apriori or that the data is missing at random or completely at random.
Depending on the missingness patterns in the target environment, different imputation strategies for missingness patterns, specifically, missing at random and missing completely at random patterns, have also been studied \citep{mayer2021generalizing}. 
However, there is a missing piece: quantification of missingness on the bias between SATE and PATE for the missing not-at-random setting, i.e., how much missingness can be tolerated to ensure transportability when covariates are not missing randomly?

Thus, while previous work has focused on the identification of covariate adjustment sets under selection bias with inconsistent or missing covariates, the identification of transported causal effects with missing covariates has not been explored to the best of our knowledge. Furthermore, quantifying the bias between SATE and PATE under selection bias is well established, but the effect of missing covariates and especially, missing mediators on the bias quantification is an open problem. In addition to addressing these research gaps, we present a framework to analyze the sensitivity to missing mediator data for estimators for the indirect effect mediated by the mediators.

%% file: methods.tex
\section{Background}

Generalizing or transporting causal effects from one observational setting consisting of a study sample in the source environment to the target environment of interest is a challenge when there are differences between the source and target environments, including those due to population distribution differences. 
Such population distribution differences across environments can arise when subgroups are underrepresented, or the treatment assignment in the target population is different from the source, in which case the estimated causal effect from the source environment cannot be directly transported to the target environment. 
Two main statistical issues result in these differences: 1) selection bias; the observational data from the source environment is a non-randomized sample of the target population; and 2) missing covariates; there is missingess in the covariates in the target environment. Formally, in the first case, if $S=1$ is an indicator of the source environment and $S=0$ is an indicator of the target environment, and $X$ is a covariate, then $P(S=1) \neq P(S=1\mid X=1)$.
The quantity of interest is the Population Average Treatment Effect (PATE) in the target environment,
\begin{align}
    \text{PATE} = E[Y^1 - Y^0 \mid S=0].
\end{align}
However, with the data in the source environment, we are only able to estimate the Sample Average Treatment Effect (SATE),
\begin{align}
    \text{SATE} = E[Y^1 - Y^0 \mid S=1].
\end{align}
PATE can be estimated using the transportability formula \citep{pearl2011transportability}. In order to estimate PATE, we consider data from two environments, $e_s$ (source environment) and $e_t$ (target environment), with treatment $A$, outcome $Y$, mediators $C$, and covariates $X$ collected in both the environments. 
Here, we assume $A$ to be a binary treatment, $A = \{0,1\}$.
In order to represent the differences between the environments that may arise due to differences in the demographics or the data measurement procedures, we use selection diagrams \citep{pearlcausality}, which include additional nodes known as selection nodes, $S$. 
These selection nodes represent the differences across environments. 
Thus, $S \rightarrow X$ represents that the only difference between the two environments is due to the distribution of $X$ \footnote{Note, an absence of an edge between $S$ and $X$ denotes that there are no differences with respect to $X$ across environments, a property which is essential to identify if the causal effects can be transported from $e_s$ to $e_t$.}. 
Next we describe the transport formula that can be used for estimating PATE.

\textbf{Causal effect estimation in the target domain.} 
In order to estimate the causal effect, $\text{PATE}$ in the target environment, $e_t$ we need to correct for the selection bias.
To do so, we represent the variables that are directly connected to $S$ by $Z$, i.e., $S \rightarrow Z$; these are the nodes that result in differences across environments.
Note that in the current framework, we only consider selection bias at one node. 
Thus, differences across environments are only attributed to a single variable. Extension of the current framework to multiple differences is left for the future. 
Accordingly, we obtain the causal effect of $A$ on $Y$ in $e_t$, $\text{PATE}$ as follows:
\begin{align} \label{eq:effect_target}
    \text{PATE} = \sum_v P(y \mid do(A=a),v_{\setminus Z}) \sum_z P^*(Z).
\end{align}
We thus have to correct for the selection bias by estimating $P^*(Z)$ in the target environment; this is the portion of the causal effect that needs to be estimated in the target domain.
Equation \ref{eq:effect_target} provides one of the ways to estimate the causal effect in the target domain even with incomplete data and selection bias.
While the transport formula helps to estimate the average treatment effect, PATE, in the target environment, estimating the indirect effect through the mediator $C$ is challenging, especially under missing data that we will discuss next.

\textbf{Mediated indirect effect estimation with missing mediator data in the target environment}
Since the goal is to understand if the indirect effect as mediated by $C$ is affected by missing data, we assume we have the treatment assignment and potential outcomes data available in the target environment as well.
Furthermore, estimating the indirect effect as mediated by $C$ can be difficult if $C$ has missing data in the target environment and there is a distribution shift in $C$ across the two environments, $S \rightarrow C$.
Specifically, we focus on the issue of where mediator data is missing in the target environment and the challenges due to the effect of missingness on estimated causal effects. Specifically, $M$ denotes the set of missingness indicators of the variables in $C$, and $\bar{C}$ denotes the set of the observed proxies of the variables in $C$ \footnote{The missing variables are represented in red in the causal graphs.}. 
The observed proxy $\bar{C}^i$ is defined as $\bar{C}^i \equiv C^{i}$ if $M_i = 1$ (i.e. the record is observed) and $\bar{C}^i = ?$ if $M_i=0$ (i.e. the record is missing). 
We illustrate the different missingness patterns with $C$ as the mediator, $A$ as the treatment, $Y$ as the outcome and $R$ as covariates through causal graphs below in Figure \ref{fig:missingness patterns} before introducing the estimator for the transported indirect effect and the sensitivity framework.

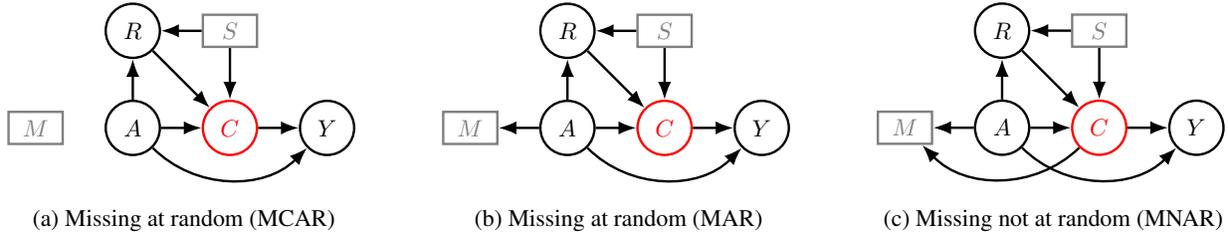
\begin{figure}[!htbp]
    \centering
    \begin{subfigure}[b]{0.3\textwidth}
    \centering
    \scalebox{0.9}{\input{Figures/mcar}}
    \caption{Missing at random (MCAR)}
    \label{fig:mcar}
    \end{subfigure}
    \hfill
    \begin{subfigure}[b]{0.3\textwidth}
    \centering
    \scalebox{0.9}{\input{Figures/mar}}
    \caption{Missing at random (MAR)}
    \label{fig:mar}
    \end{subfigure}
    \hfill
    \begin{subfigure}[b]{0.3\textwidth}
    \centering
    \scalebox{0.9}{\input{Figures/mnar}}
    \caption{Missing not at random (MNAR)}
    \label{fig:mnar}        
    \end{subfigure}
    \caption{Different missingness patterns for the mediator $C$ illustrated using causal graph for missing completely at random (MCAR) in (a) where $C \independent M$, missing at random  (MAR) in (b) where $C \independent M \mid A, R$, and missing not at random (MNAR) in (c) where $C \not\!\perp\!\!\!\perp M \mid A,R$.}
    \label{fig:missingness patterns}
\end{figure}

These three settings, MCAR, MAR, and MNAR represent the different missingness patterns commonly found in real-world datasets. For assessing the effect of the missing mediator data, we focus especially on the MNAR setup as illustrated next as mediator data, especially about individual behavior, is found to be missing, not at random \citep{hallgren2013missing,hallgren2016missing}. Moreover, in this particular case, correcting for the bias due to missingness is challenging as opposed to MCAR and MAR \citep{wang2010correction}.  

\section{Sensitivity of the causal estimator to missingness of mediators.}
\begin{figure}[!htbp]
    \centering
    \scalebox{1.2}{\input{Figures/step_5}}
    \caption{Example figure representing the interaction between neighborhood factors and CVD risk; neighborhood SES ($A$) affects individual behaviors related to CVD, such as alcohol consumption, which are mediators ($C$), while $C$ also affects CVD risk. $A$ also affects the alcohol resources in the neighborhood, $R$ which in turn affects $C$. Alcohol consumption and CVD risk can be confounded by `race', $W$, which also defines disadvantaged and advantaged groups based on the value of `race'. Behavioral data for individuals related to alcohol consumption, the mediators can be missing for specific subgroups \citep{sania2021k}, which is denoted by the missingness indicator ($M$). We also observe a distribution shift with respect to the individual behavior related to alcohol consumption which is represented by the selection node ($S$).}
    \label{fig:example_ph}
\end{figure}
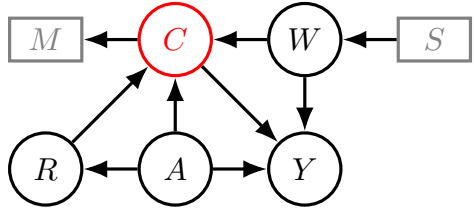

Next, we present a specific case of estimating the causal effect with missing mediators, especially when the missingness is `missing not at random.' Figure \ref{fig:example_ph} represents one such setup. This is the causal graph we will be using for simulated analysis in this study. The causal graph is postulated based on findings from public health studies where factors at the neighborhood level, such as neighborhood socioeconomic status, affect individual behaviors such as alcohol consumption and also affect individual health outcomes \citep{skagerstrom2011predictors,sania2021k,simkhada2008factors}.
\subsection{Estimator in the target environment with missing data and selection bias}
We assume that in the target domain, the covariates that are missing are also the ones that account for the distribution shift. Thus, we have the following edges in the DAG, $S \rightarrow W \rightarrow C$. 
\cut{
Next we represent the potential outcome model where for simplicity, we consider a additive linear model as follows:
\begin{align}
    \ep[Y^a] =  \beta_a A + \beta_{ac} C + \beta_{ar} R 
\end{align}
Thus, the treatment effect for individual $i$ is
\begin{align}
    \ep[Y^1_i] - \ep[Y^0_i] =  \beta_a + \beta_{ac} C_i + \beta_{ar} R_i 
\end{align}}
Following \citep{rudolph2021transporting}, we represent the nontransported average treatment effect as follows:
\begin{align}
    \text{SATE} := \Psi \left(P_{S=1}\right) =  \ep\left[Y_{a,g_{C\mid a^*,S=1}} \right],
\end{align}
where the expectation is taken over the full data model and $Y_{a,g_{C\mid a^*}}$ is a potential outcome intervening on $A$ to set it to a specific treatment $a$, and then intervening on $C$ to set it to a random draw from the distribution of $C$,i.e., a stochastic intervention and $g^*_{C\mid a^*,s, w}$ captures this intervention effect on $C$, where the distribution is defined as follows:
\begin{align}
g^*_{C\mid a^*,s, w} = \sum_r P\left(C=c \mid A=a^*, R=r, S=s, W\right)  \times P\left(R=r\mid A=a^*\right) .
\end{align}

When the effect is transported to the target environment (S=0), we make the following modification:
\begin{align}
    \text{TATE} := \Psi(P_{S=0}) =  \ep\left[Y_{a,g^*_{C\mid a^*,w, S=0}} \right],
\end{align}
where we have to adjust for the mediators in the target environment, $g_{C\mid a^*, w S=s} = \sum_r P\left(C=c \mid A=a^*, R=r, S=0, W\right)\times P\left(R=r\mid A=a^*, S=0\right)$.
The transported stochastic direct effect (SDE) can be evaluated by setting $a^*$ to 0 and taking the mean difference in the outcome between setting $a$ to 1 and setting $a$ to 0,
\begin{align}
    \text{SDE} = \ep \left[Y_{1,g_{C\mid 0,s,w}} - Y_{0,g_{C\mid 0,s,w}} \mid S=0 \right].
\end{align}
and the transported stochastic indirect effect  (SIE)can be evaluated by setting $a=1$ and then taking the difference in the mean outcome between setting $a^* = 1$ and $a^*=0$, denoted as:
\begin{align}
    \text{SIE} = \ep \left[Y_{1,g_{C\mid 1,s,w}} - Y_{1,g_{C\mid 0,s,w}} \mid S=0 \right].
\end{align}

Since only SIE depends on the effect of the mediator, represented by $g_{C\mid 0, S=s, W}$, we focus on the SIE for sensitivity to the missingness of the mediator. 

Moreover, since we are interested in assessing how SIE would vary across the missingness of the mediator for different subgroups of the data, we focus on $\text{SIE}_{W=0}$ to represent the stochastic indirect effect in the target environment for the disadvantaged group and compare this against $\text{SIE}_{W=1}$, the stochastic indirect effect in the target environment for the advantaged group. The conditional stochastic effect for a specific value of $W=w*$ is denoted as follows:

\begin{align}
    \text{SIE} = \ep \left[Y_{1,g_{C\mid 1,s,w=w*}} - Y_{1,g_{C\mid 0,s,w=w*}} \mid S=0, W=w* \right].
\end{align}

\subsection{Targeted minimum loss-based estimator (TMLE) for the target environment}
We focus on the TMLE estimator proposed by \citet{rudolph2021transporting} while adapting it for the setting where the treatment has a direct effect on the outcome.
In order to estimate the parameter of interest, $\Psi \left(P_{S=s}\right)$, we write the efficient influence curve equation of the parameter (EIC). 
An EIC is defined in terms of the statistical model, $M$ defined by the causal graph in Figure \ref{fig:mcar} and the target parameter, $\Psi (P_{S=s})$, as the canonical gradient of the pathwise derivative of the $\Psi (P_{S=s})$ along each possible submodel of $M$. 
If a regular asymptotically linear (RAL) estimator has an influence curve equal to EIC, then the estimator is asymptotically efficient, meaning it is of minimum variance for an RAL estimator. 
Accordingly, the empirical mean of the influence curve provides a linear approximation of the estimator. Thus, the empirical mean of the influence curve is the parameter of interest here. 

The EIC of this parameter is given by
\begin{align}
\begin{split}
        D(P) &= D_Y(P) + D_R(P), \text{where} \\
        D_Y(P) &= \left(Y - \bar{Q}_Y\left(A,C,R\right)\right) \times \frac{g^*_{C,n\mid a^*,s}(C)P_R\left(R\mid A=a,S=0\right)I\left(S=1\right)}{g_C(C\mid R,S=1)P_R\left(R\mid S=1\right)P_S(S=0)} \\
        D_R(P) &= \left(\bar{Q}_C\left(a,R,S\right) - \Psi (P)\right)\frac{I(S=0)}{P_S(S=0)}.
\end{split}
\end{align}
Let $\bar{Q}^0_{Y,n}(R,C,A)$ be an initial estimate of $\ep \left[Y \mid C,R,A\right]$. $\bar{Q}^0_{Y,n}(R, C, A)$ can be estimated by predicted values from a regression of Y on $C, R, A$ among those with $S=1$ in the source environment. We update the initial estimate of $\bar{Q}^0_{Y,n}(R,C,A)$ using the following weights,
\begin{align}
    H (C, R, A, S) = \frac{g^*_{C,n\mid a^*,s}(C)P_R\left(R\mid A=a,S=0\right)I\left(S=1,A=a\right)}{P_C\left(C\mid R,S=1\right)P_R\left(C\mid A=a, S=1\right)P_A(a\mid S=1) P_S(S=0)}
\end{align}
$\bar{Q}^0_{Y,n}(R,C,A)$  is then updated by performing a weighted logistic regression of $Y$ with $\bar{Q}^0_{Y,n}(R,C,A)$  as an offset and intercept $\epsilon_Y$ and weights $H_n\left(C,R,A,S\right)$. The update is given by $\bar{Q}^*_{Y,n}(R,C,A)$ = $\bar{Q}^0_{Y,n}(\epsilon_{Y,n})(R,C,A)$.

We can then perform the stochastic intervention on $\bar{Q}^*_{Y,n}(R,C,A)$ via the computation $\bar{Q}^*_{C,n}(R,A,C) = \ep_{g^*_{C,n\mid a^*,s}}\left[\bar{Q}^*_{Y,n}(R,C,A) \mid A,S\right]$. This can be done by generating predicted values of $\bar{Q}^*_{Y,n}(R,1,A)$ and $\bar{Q}^*_{Y,n}(R,0,A)$ and then marginalizing over $g^*_{C,n\mid a^*,s}(C): \sum_{c=0}^1 \bar{Q}^*_{Y,n}(R,c,A)g^*_{C,n\mid a^*,s}(c)$.

The empirical mean of $\bar{Q}^*_{C,n}(R, A, C)$ among those for whom $S=0$ is the TMLE estimate of $\psi(P)$. It solves $\frac{1}{n}\sum_i=0 ^n D^*_n(P) = 0$. The transported stochastic direct effect (SDE) can be evaluated by setting $a^*$ to 0 and taking the difference in $\psi(P)$ setting $a$ to 1 versus setting $a$ to 0. The transported stochastic indirect effect (SIE) entails setting $a=1$ and then taking the difference in $\psi (P)$ setting $a^* = 1$ versus $a^* = 0$. The corresponding EIC is the difference in EIC for the parameter defined by setting $a^*=1, a=1$ and the EIC for the parameter defined by setting $a^*=0,a=1$.

Since the empirical mean of $\bar{Q}^*_{C,n}(R, A, C)$ among those for whom $S=0$ is the TMLE estimate, and the expectation is taken over $g^*_{C,n\mid a^*,s}(c)$, having missing values in $C$ will influence the effect. Derive how the effect will look for indirect effect and get it in terms of $g^*_{C,n\mid a^*,s}(c)$.

In this analysis, when updating the initial estimates while performing MLE, we adjust for the target-specific factors that will be affected by the missingness in $C$. We next describe how the specific weights that are affected by missingness, which involves the mediators in the target distribution - 
\begin{align}
    \mathbf{w}^* = g^*_{C,n\mid a^*,s}(C).
\end{align}
Here, if there is missingness in $C$, the estimate of $\mathbf{w}^*$ will be biased. We denote the biased estimate by $\bar{\mathbf{w}} = g^*_{C^*,n\mid a^*,s}(C^*)$, where $C^*$ denotes $C$ without the missing data in $C$, that is we only select the samples for which $C$ is not missing.
\cut{In order to transport the effect from the source to the target environment, we need to estimate the potential outcome by performing a weighted parametric regression where the weights for the parameterization as follows:
\begin{align}
    w = \frac{g_{C\mid a^*,S=0}(C\mid A,S=0)}{p_C(C\mid A,S=1) p_A(a\mid S=1)p_S(S=1)}
\end{align}

However, when we have missing data for the mediator, the weights for the weighted parametric regression, $w*$ would be biased to missingness with missing data for the mediator represented by $C*$,
\begin{align}
    w* := \frac{g_{C*\mid a^*,S=0}(C*\mid A,S=0)}{p_C(C\mid A,S=1) p_A(a\mid S=1)p_S(S=1)}
\end{align}
}
\subsection{Sensitivity framework}
The parametric formulation of the weighted regression is sensitive to the missing data of the mediator. In order to assess the sensitivity, we focus on the classical approach by \citet{tan2006distributional} specific to the weights for individuals where $ i = 1, \dots n$, as follows:
\begin{align}
    \lambda^{-1} \leq \frac{\mathbf{w}_i*}{\mathbf{w}_i}\leq \lambda,
\end{align}
where $\lambda \geq 1$. 

For the indirect effect in the target environment, we consider the effect on $g_{C*\mid a^*,S=0}(C*\mid A,S=0)$ with $k$ missing data points as follows:

\begin{definition}
Let $R^2$ be the residual variation in the true weights $w^*$ , not explained by the estimated weights $\bar{w}$ with missingness:
\begin{align}
    R^2  := 1- \frac{var(\bar{\mathbf{w}}_{-(i)}\mid i\in k)}{var(w_i^*)}
\end{align}
Then for a fixed $R^2 \in [0,1)$, we define the set of variance-based sensitivity models as $\sigma(R^2)$:
\begin{align}
    \sigma(R^2) = \left\{ \mathbf{w}_i^* \in \mathbb{R}^n : 1 \leq \frac{var(\mathbf{w}_i^*)}{var(\bar{\mathbf{w}}_{-(i)}\mid i\in k)} \leq \frac{1}{1-R^2} \right\}
\end{align}

\end{definition}
Accordingly, the estimate of $\text{SIE}^*$ will be biased function of $R^2$. We represent $\text{SIE}^*$ as follows:
\begin{align}
\begin{split}
     \text{SIE}^* &= \ep_{g^*_{C,n\mid a^*=1,s=0}}\left[\bar{Q}^*_{Y,n}(R,C,A=1) \mid A=1 ,S=0\right] \\
     &- \ep_{g^*_{C,n\mid a^*=0,s=0}}\left[\bar{Q}^*_{Y,n}(R,C,A=1) \mid A=1 ,S=0\right]\\
     &:= \delta(R^2) \text{SIE},
\end{split}
\end{align}
where $\delta(R^2)$ is a function of the variance that is not explained by the missing weights, $\bar{w}_{-{i}}$. 

Applying the approach from \citet{huang2022variance}, we derive the confidence intervals for the $\text{SIE}^*$ as follows:
\begin{align}
    \text{CI}(\alpha) = \left[ D_{\alpha/2}\left(\text{inf}_{\bar{w}\in R^2} \text{SIE}^*\left(\delta\left(R^2\right)\right)\right),D_{1-\alpha/2}\left(\text{sup}_{\bar{w}\in R^2} \text{SIE}^*\left(\delta\left(R^2\right)\right)\right) \right]
\end{align}

%% file: Figures/mcar.tex



\begin{tikzpicture}
    \node[state] (a) at (0,0) {$A$};
    \node[state] (r) [above =of a] {$R$};
    \node[state, red] (p) [ right =of a] {$C$};
    \node[state] (y) [right =of p] {$Y$};
    \node[state,rectangle,gray] (s) [right =of r] {$S$};
    \node[state,rectangle,gray] (m) [left =of a] {$M$};

    \path (a) edge (p);
    \path (a) edge (r);
    \path (r) edge (p);
    \path (s) edge (p);
    \path (s) edge (r);
    \path (p) edge (y);
    \path (a) [bend right=45] edge (y);

\end{tikzpicture}

%% file: Figures/mar.tex



\begin{tikzpicture}
    \node[state] (a) at (0,0) {$A$};
    \node[state] (r) [above =of a] {$R$};
    \node[state, red] (p) [ right =of a] {$C$};
    \node[state] (y) [right =of p] {$Y$};
    \node[state,rectangle,gray] (s) [right =of r] {$S$};
    \node[state,rectangle,gray] (m) [left =of a] {$M$};

    \path (a) edge (p);
    \path (a) edge (r);
    \path (r) edge (p);
    \path (s) edge (p);
    \path (s) edge (r);
    \path (a) edge (m);
    \path (p) edge (y);
    \path (a) [bend right=45] edge (y);

\end{tikzpicture}

%% file: Figures/mnar.tex



\begin{tikzpicture}
    \node[state] (a) at (0,0) {$A$};
    \node[state] (r) [above =of a] {$R$};
    \node[state, red] (p) [ right =of a] {$C$};
    \node[state] (y) [right =of p] {$Y$};
    \node[state,rectangle,gray] (s) [right =of r] {$S$};
    \node[state,rectangle,gray] (m) [left =of a] {$M$};

    \path (a) edge (p);
    \path (a) edge (r);
    \path (r) edge (p);
    \path (s) edge (p);
    \path (s) edge (r);
    \path (a) edge (m);
    \path (p) edge (y);
    \path (a) [bend right=45] edge (y);
    \path (p) [bend left=45] edge (m);

\end{tikzpicture}

%% file: Figures/step_5.tex
\begin{tikzpicture}
    \node[state] (a) at (0,0) {$A$};
    \node[state,red] (p) [ above  =of a] {$C$};
    \node[state] (y) [  right =of a] {$Y$};
    \node[state] (r) [left =of a]{$R$};
    \node[state] (w) [right =of p] {$W$};
    \node[state,rectangle,gray] (s) [right =of w] {$S$};
    \node[state,rectangle,gray] (m) [left =of p] {$M$};

    \path (a) edge (p);
    \path (p) edge (y);
    \path (s) edge (w);
    \path (a) edge (y);
    \path (a) edge (r);
    \path (r) edge  (p);
    \path (p) edge (m);
    \path (w) edge (p);
    \path (w) edge (y);

\end{tikzpicture}

%% file: experiments.tex
\section{Simulation Study}
\subsection{Data and Variables}
Based on the causal graph presented in Figure \ref{fig:example_ph}, we sample data from a distribution as follows:
\begin{align*}
        &A \sim \text{Bernoulli}\left(0.5\right) \\
        &W \sim \text{Bernoulli}\left(0.5*S + \cn\left(0,0.1\right)\right) \\
        &R := 0.7* A + \cn\left(0, 0.5\right)\\
        &C := 1.5* R + 0.2* W + 0.8* (1-W) + \cn\left(0, 0.5\right)\\
        &Y = \text{Bernoulli}\left(\sigma\left(0.2*A + 2.5*C+ -0.7*W\right)\right)\\
        &M := \text{Bernoulli}\left(\sigma\left(\lambda*C*W + 0*C*(1-W)\right)\right)\\
        &\sigma(x)=1/(1+\text{exp}(-x))
    \end{align*}
Here, $S=1$ denotes the target environment, $W=1$ denotes the advantaged group with complete mediator data without missingness, $W=0$ denotes the disadvantaged group with missing mediator data, and $\lambda$ represents the proportion of missingness. The specific parameters for the data-generating process were chosen to allow for the indirect effect of $A$ on $Y$ to be conditionally the same given $W$ for the advantaged ($W=1$) and disadvantaged groups ($W=0$).
Following the data-generating process in accordance with the causal graph, we varied the proportion of missingness of the mediator $C$ from 0.1 to 0.9 for the disadvantaged group, thus affecting the bias in $R^2$ for the disadvantaged group while not affecting the advantaged group.

\begin{figure}[!htbp]
    \centering
    \includegraphics[scale=0.4]{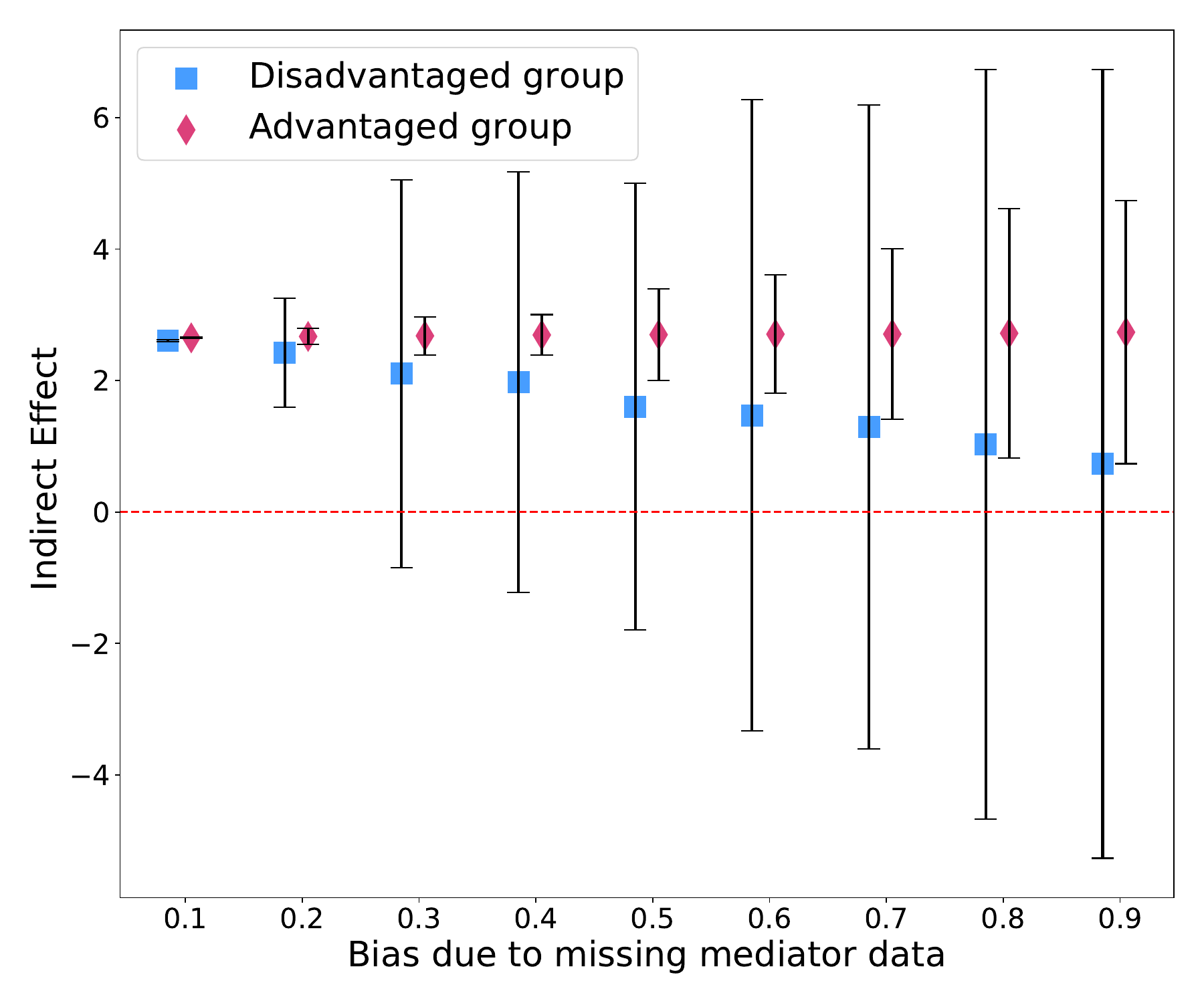}
    \caption{Results from the sensitivity analysis under the sensitivity framework. We vary the bias in $R^2$ measure due to missing mediator data across the x-axis for the disadvantaged and advantaged groups independently and plot the range of estimated stochastic indirect effect (SIE) values on the y-axis. The solid bar denotes the point estimate bounds for a specified bias in $R^2$ value, estimated as the point estimate plus 95\% confidence intervals. We estimate a bias in $R^2$ = 0.29, such that if $R^2 \geq 0.29$, the intervals contain the null estimate for the SIE.    }
    \label{fig:sensitivity}
\end{figure}

\subsection{Results}
While the stochastic indirect effect is homogenous across advantaged and disadvantaged groups without any missingness in $C$, increasing the bias in $R^2$ makes the indirect effect heterogeneous when compared between the advantaged and disadvantaged group, as illustrated in Figure \ref{fig:sensitivity}.
Moreover, the indirect effect, as mediated by $C$, becomes insignificant for bias in $R^2 > 0.29$ for the disadvantaged group, which has missing data for $C$, while it remains significant for the advantaged group, which does not have missing data for $C$.
The indirect effect for the disadvantaged group for a bias in $R^2$ of 0.9 is approximately 0.31 with 95\% confidence interval of (-5.20, 6.51), thus being insignificant as compared with a bias in $R^2$ of 0.1 with the indirect effect mediated by $C$ approximately equal to 2.72 with the 95\% confidence interval of (2.62, 2.81) which is significant.
The indirect effect for the advantaged group for a bias in $R^2$ of 0.9 is, on the contrary, approximately equal to 2.79 with the 95\% confidence interval of (0.41, 4.41) and thus remains significant with an increasing bias in $R^2$.

\section{Application of Sensitivity Framework to Moving to Opportunity Study}
\subsection{Background}
Moving to Opportunity (MTO) was a longitudinal randomized trial conducted by the U.S. Department of Housing and Urban Development from 1994 to 2007 in five cities in the United States: Baltimore, Boston, Chicago, Los Angeles, and New York \citep{sanbonmatsu2011moving}. 
Families living in high-rise public housing in these cities could sign up to be randomized at baseline, the starting point of the experiment, to receive a Section 8 housing voucher that they could use to move out of public housing and into rental housing on the private market.
The adult participants and their children were then surveyed at two follow-up time points.
The goal was to estimate the effects of the housing intervention on economic, educational, and health outcomes of adults and children over time. 
The MTO study provides an opportunity to uncover how improving the neighborhood improves individual-level health outcomes.
Moreover, since this is a multicity dataset, we can assess the transportability of interventions across environments, specifically cities.
Previous work has also used this dataset to assess the transportability of interventions while not focusing on issues related to missing data \citep{rudolph2021transporting}. This motivates the use of this dataset to assess further the transportability challenges under missing data. 

\subsection{Data and Variables}
In this work, we specifically focus on receiving a Section 8 housing voucher as treatment, $A$, and child mental health or substance abuse during the follow-up years as the outcome $Y$. 
The specific mediator we focus on is overall parental health, $C$, which is known to mediate the effect of neighborhood factors on youth substance abuse \citep{buu2009parent}. 
We incorporated an extensive set of covariates, $W$, at the individual and family levels, including sociodemographics, neighborhood characteristics as reported by the adult family member, and reasons for participation that were controlled for in the analysis.
Based on prior research with  MTO \citep{rudolph2018composition}, we consider New York, Los Angeles, Boston, and Chicago as the target environments.
Data from Baltimore was not considered in the analysis since voucher receipt was not associated with moving to a lower-poverty neighborhood, unlike other cities \citep{rudolph2018composition}.
While data from one city was used as the target environment, data from the rest of the three cities was combined as the source environment. For example, with New York as the target environment, data from Los Angeles, Boston, and Chicago was combined as the source environment.
To assess the sensitivity to missingness, we induce missingness in the mediator, `parental health’ for the disadvantaged group determined based on self-reported race by individuals. This allows us to assess under what proportions of missingness the causal effect of receiving a Section 8 voucher becomes insignificant.
Missingness is introduced randomly in overall parental health, considered as the mediator, $C$ for 0.1, 0.2, 0.3, 0.4, and 0.5 proportion of the total sample for the disadvantaged group.
Moreover, since African Americans and Latinos comprised the majority of the racial and ethnic groups, we considered African Americans as the disadvantaged group ($W=0$) and Latinos as the advantaged group ($W=1$). This follows from previous studies that have focused on African American and Latino subgroups for analysis in MTO \citep{rudolph2018composition}.


\begin{figure}[!htbp]
    \centering
    \begin{subfigure}[b]{0.45\textwidth}
        \centering
        \includegraphics[scale=0.27]{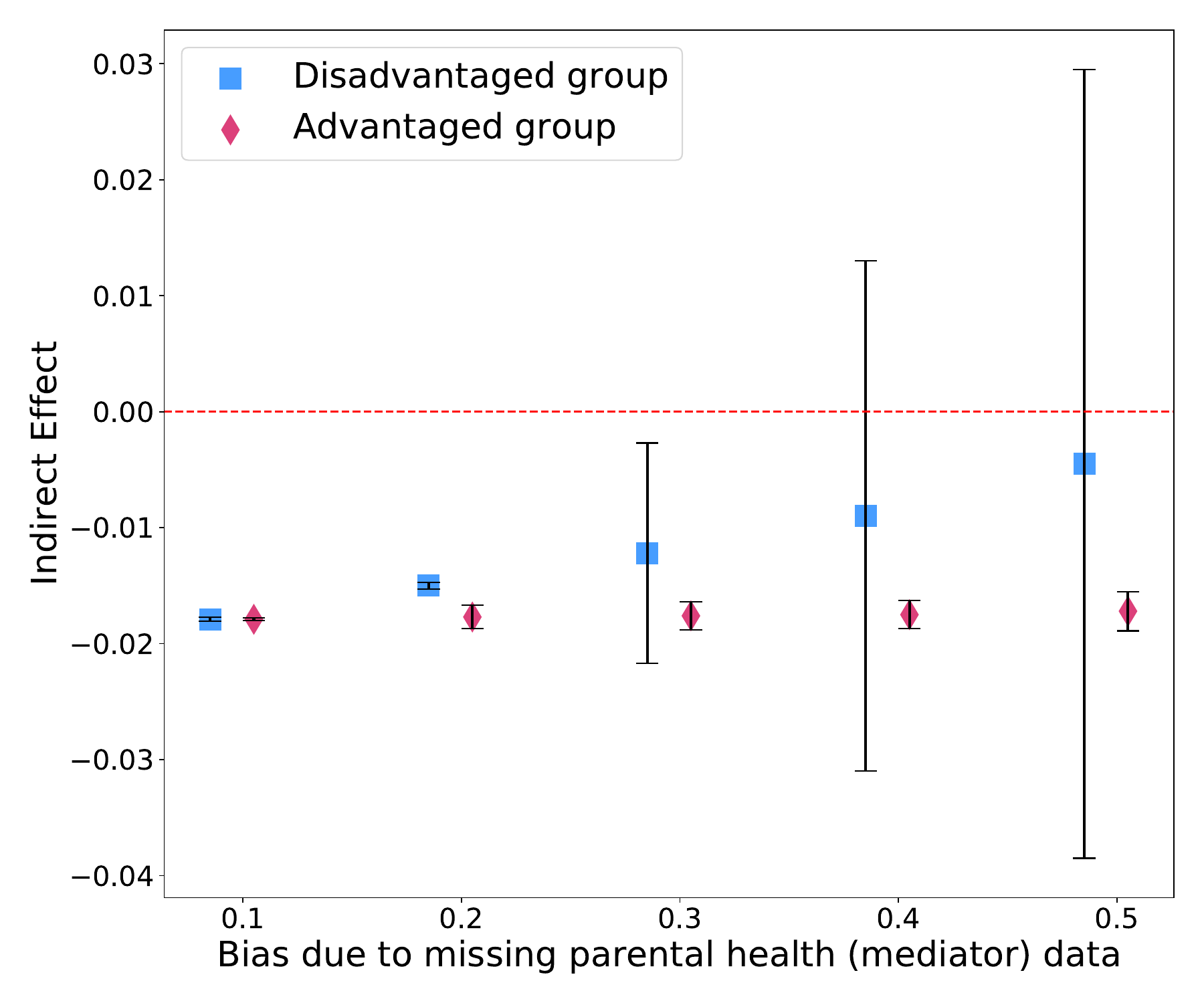}
        \caption{}
    \end{subfigure}
     \begin{subfigure}[b]{0.45\textwidth}
        \centering
        \includegraphics[scale=0.27]{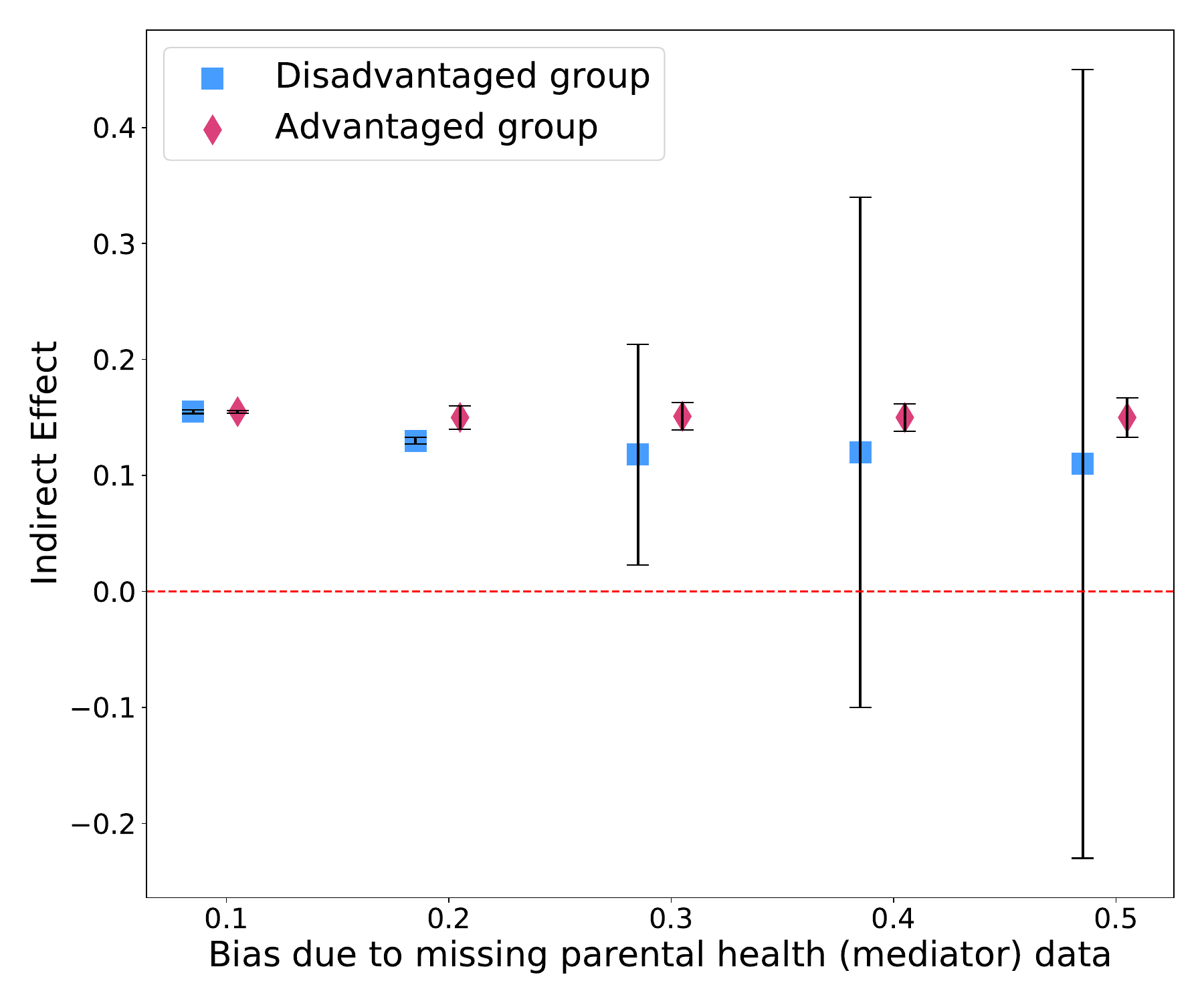}
        \caption{}
    \end{subfigure}
         \begin{subfigure}[b]{0.45\textwidth}
        \centering
        \includegraphics[scale=0.27]{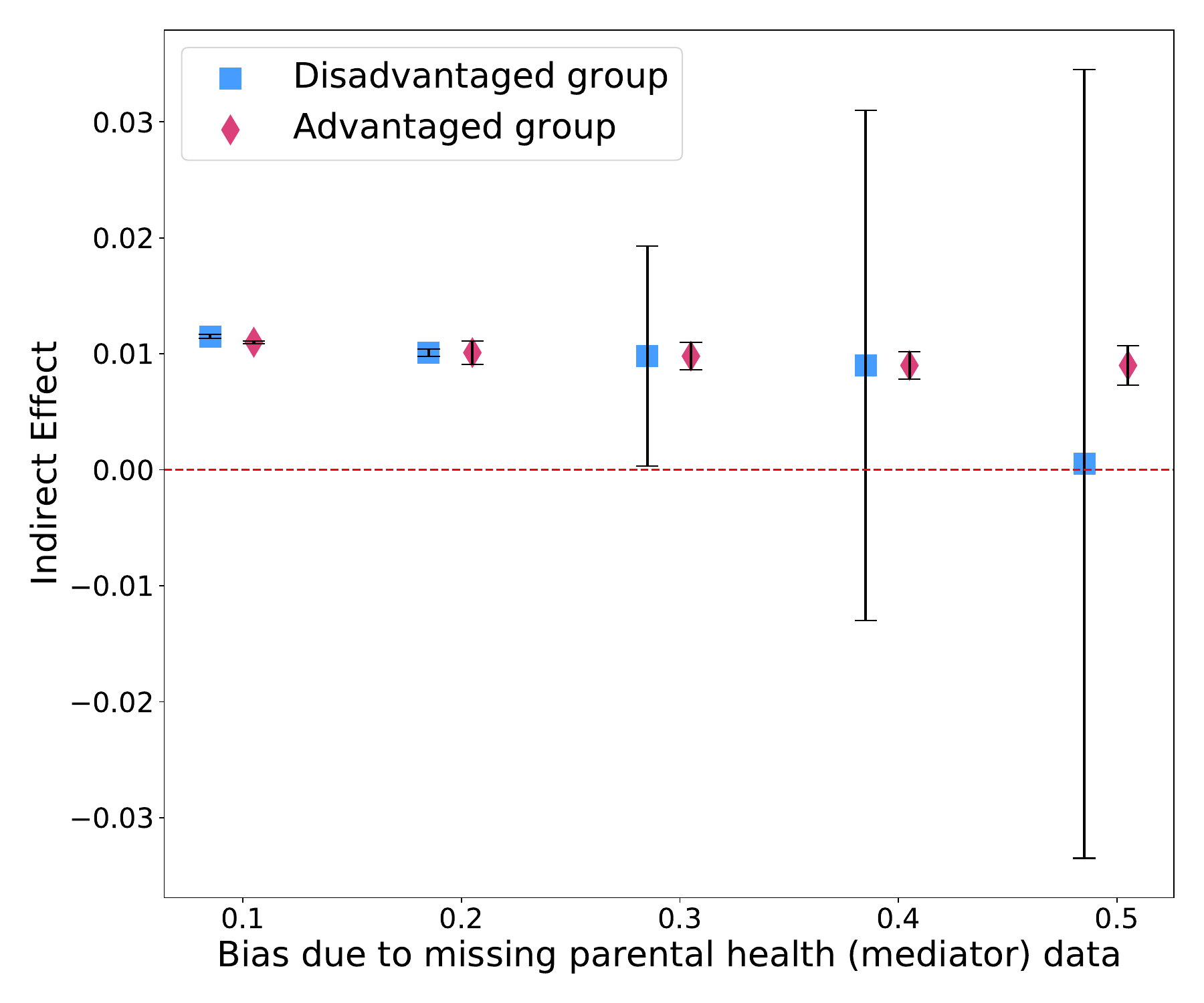}
        \caption{}
    \end{subfigure}
         \begin{subfigure}[b]{0.45\textwidth}
        \centering
        \includegraphics[scale=0.27]{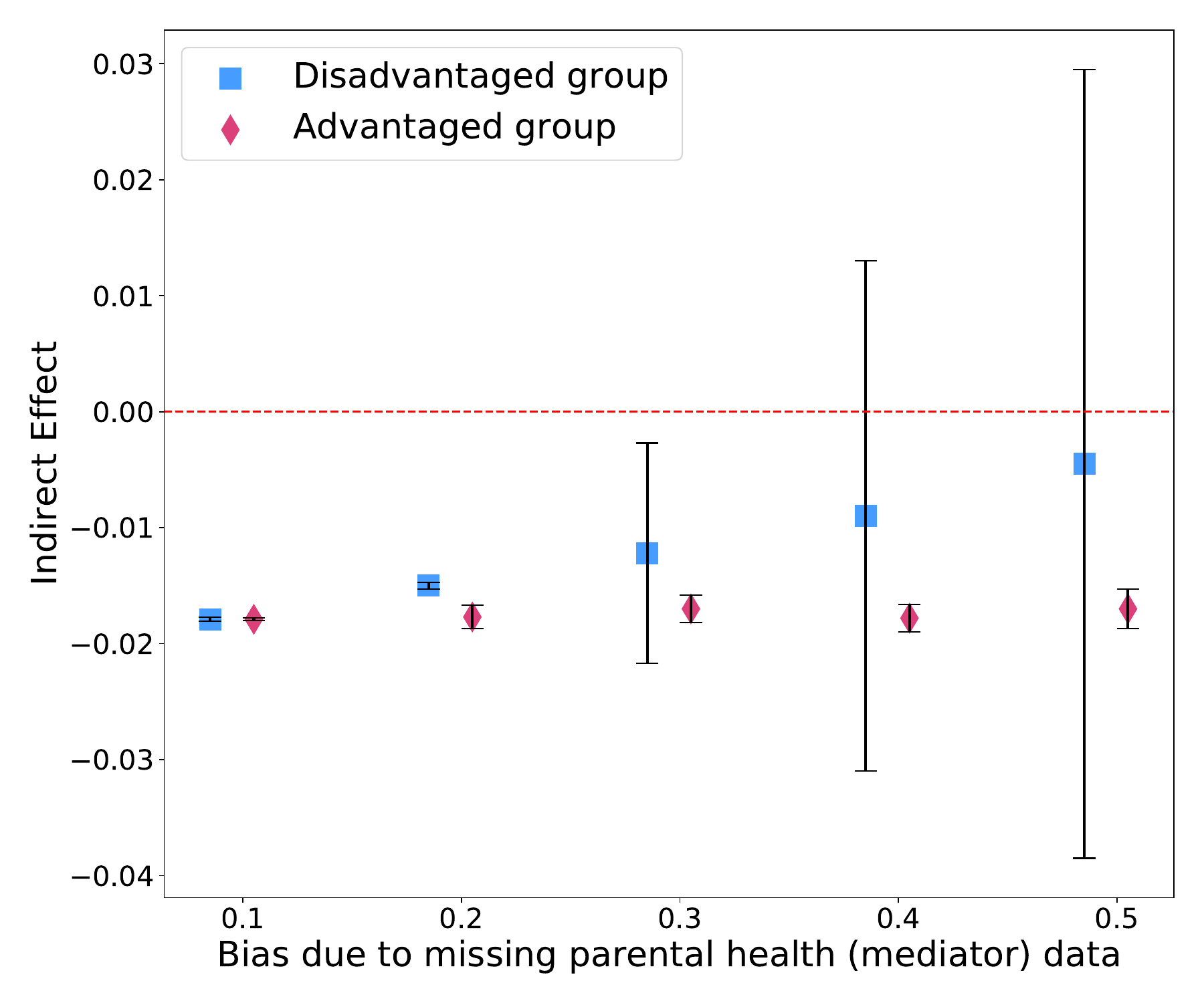}
        \caption{}
    \end{subfigure}
    \caption{Results from the sensitivity analysis under the sensitivity framework for MTO data with Los Angeles as the target environment in (a), Boston as the target environment in (b), New York City as the target environment in (c), and Chicago as the target environment in (d). We vary the bias in $R^2$ due to missing mediator (parental health) across the x-axis for the minority and majority groups independently and plot the range of estimated stochastic indirect effect (SIE) values on the y-axis. The solid bar denotes the point estimate bounds for a specified proportion of missing mediator data, estimated as the point estimate plus 95\% confidence intervals.}
    \label{fig:sensitivity_mto}
\end{figure}

\subsection{Results}
We assess the point estimate of the stochastic indirect effect along with the confidence interval as a function of the variance in the weights not explained by the missing mediator data, $R^2$.
With a missingness proportion close to 0.1, the point estimate is approximately  -0.018 for Los Angeles, 0.161 for Boston, 0.012 for New York, and -0.018 for Chicago for the advantaged and disadvantaged groups as shown in Figure \ref{fig:sensitivity_mto}.
This effect direction aligns with previous research, which found that moving to lower-poverty neighborhoods reduces overall substance abuse and improves mental health for youth in certain cities such as Los Angeles and Chicago \citep{sanbonmatsu2011moving} but not in the case of New York and Boston.
However, as missingness increases, the indirect transported causal effect estimate is close to -0.009 for the disadvantaged group with confidence intervals (-0.03, 0.015) for a bias around 0.37, thus suggesting that the effect becomes insignificant for the disadvantaged group for Los Angeles. 
On the other hand, for the advantaged group, the stochastic indirect effect is close to -0.192 with a confidence interval (-0.182, -0.206) and is thus significant for the advantaged group. 
As bias due to missingness increases to 0.5, the indirect effect in Los Angeles for the disadvantaged group is close to -0.002 with a confidence interval (-0.044, 0.031), and for the advantaged group is approximately -0.191 with a confidence interval (-0.182, -0.206). We also observe similar effects in the case of Chicago.
However, in the case of Boston, with increasing missingness, the stochastic indirect effect for the disadvantaged group is 0.101 with a confidence interval (-0.210,0.422) while that for the advantaged group is 0.160 with a confidence interval (0.159,0.161) for a bias of 0.5 due to missing parental data (mediator). In the case of New York, for a 0.5 bias due to missingness, the indirect effect for the disadvantaged group is 0.001 with a confidence interval (-0.032,0.033) for the disadvantaged group.
As the bias due to missingness increases, we find that the transported indirect effect for Los Angles, Boston, New York, and Chicago becomes insignificant for the disadvantaged group but remains significant for the advantaged group.

%% file: discussion.tex
\section{Discussion}\label{sec:discussion}
In this work, we focus on the issue of transporting the causal effects of interventions from a source environment to a target environment when data is missing in the target environment.
Specifically, we study the situation where mediator data is missing in the target environment. 
This is an important problem because of many situations in which it has been demonstrated that data about mediators, such as individual behaviors like smoking status or alcohol consumption, are inconsistently obtained, resulting in missing data.
Since these individual behaviors mediate the effect of population-level factors such as neighborhood socioeconomic status on health outcomes, our study examines to what degree their missingness can introduce bias in the estimated indirect causal effect for the disadvantaged group compared to the advantaged group.
Our contributions include introducing a novel set of models for assessing the sensitivity of a transported indirect causal effect to missing mediator data. We show how the variance-based sensitivity models can be parameterized with respect to an $R^2$ measure representing the degree of residual error due to an omitted mediator.
In simulated data, analyzing the residual bias in the estimated transported effect by varying the magnitude of the missingness shows that the transported causal effect becomes insignificant for the disadvantaged group as the missingness increases beyond a threshold of 0.29 for our causal graph. 
We observe similar characteristics when evaluating the effect of moving to a better neighborhood on child mental health and substance abuse where there is also missingness in the mediator, overall parental health; missingness in the parental health renders the effect of moving to a better neighborhood on child mental health and substance abuse insignificant for the disadvantaged group.\\

Certain limitations to this framework should be noted. Our approach relies on the assumption that the mediator's missingness pattern in the target environment is known; adopting the framework where this assumption fails is a potential future direction.
Moreover, while the analysis in this study was performed by only including the samples for which data is not missing, adapting different missing data imputation strategies to assess the bias in the transported causal effect is left for the future. 
This is challenging, considering that the missingness pattern we are concerned with is specific to `missing not at random,' which can involve potential bias depending on the imputation strategy.
Moreover, considering settings where there could be unmeasured confounding between the treatment and the outcome, the treatment and the mediator, and the mediator and the outcome is another interesting future direction.